\documentclass[10pt,twocolumn,letterpaper]{article}

\usepackage{iccv}
\usepackage{times}
\usepackage{epsfig}
\usepackage{graphicx}
\usepackage{amsmath}
\usepackage{amssymb}

\usepackage[pagebackref=true,breaklinks=true,letterpaper=true,colorlinks,bookmarks=false]{hyperref}

\iccvfinalcopy % *** Uncomment this line for the final submission

\DeclareMathOperator*{\argmin}{arg\,min}

\newcommand\T{{\hspace{-1pt}\intercal}}

\def\figvspace{{\vspace{-4mm}}}
\newcommand{\Paragraph}[1]{\vspace{-0mm} \noindent \textbf{#1} \hspace{0mm}}

\def\iccvPaperID{6} % *** Enter the ICCV Paper ID here

\ificcvfinal\fi
\begin{document}

%%%%%%%%% TITLE
\title{Dense Face Alignment}

\author{Yaojie Liu$^1$, Amin Jourabloo$^1$, William Ren$^2$, and Xiaoming Liu$^1$\\
$^1$Department of Computer Science and Engineering, Michigan State University, MI\\
$^2$Monta Vista High School, Cupertino, CA\\
$^1${\tt\small $\{$liuyaoj1,jourablo,liuxm$\}$@msu.edu},
$^2${\tt\small williamyren@gmail.com }
}

\maketitle
%\thispagestyle{empty}

%%%%%%%%% ABSTRACT
\begin{abstract}
Face alignment is a classic problem in the computer vision field. 
Previous works mostly focus on sparse alignment with a limited number of facial landmark points, i.e., facial landmark detection. 
In this paper, for the first time, we aim at providing a very dense $3$D alignment for large-pose face images. 
To achieve this, we train a CNN to estimate the $3$D face shape, which not only aligns limited facial landmarks but also fits face contours and SIFT feature points. 
Moreover, we also address the bottleneck of training CNN with multiple datasets, due to different landmark markups on different datasets, such as $5$, $34$, $68$.
Experimental results show our method not only provides high-quality, dense $3$D face fitting but also outperforms the state-of-the-art facial landmark detection methods on the challenging datasets. Our model can run at real time during testing and it's available at \url{http:///cvlab.cse.msu.edu/project-pifa.html}.
\end{abstract}

\section{Introduction}
Face alignment is a long-standing problem in the computer vision field, which is the process of aligning facial components, e.g., eye, nose, mouth, and contour. 
An accurate face alignment is an essential prerequisite for many face related tasks, such as face recognition~\cite{ding2016multi}, $3$D face reconstruction~\cite{7776921,unconstrained-3d-face-reconstruction} and face animation~\cite{zhou2016method}. 
There are fruitful previous works on face alignment, which can be categorized as generative methods such as the early Active Shape Model~\cite{milborrow2008locating} and Active Appearance Model (AAM) based approaches~\cite{kossaifi2016fast}, and discriminative methods such as regression-based approaches\cite{zhu2015face,xiao2016robust}. 

Most previous methods estimate a {\it sparse} set of landmarks, e.g., $68$ landmarks.  
As this field is being developed, we believe that Dense Face Alignment (DeFA) becomes highly desired.
Here, DeFA denotes that it's doable to map any face-region pixel to the pixel in other face images, which has the {\it same} anatomical position in human faces.
For example, given two face images from the same individual but with different poses, lightings or expressions, a perfect DeFA can even predict the mole (i.e. darker pigment) on two faces as the same position.
Moreover, DeFA should offer dense correspondence not only between two face images, but also between the face image and the canonical $3$D face model. 
This level of detailed geometry interpretation of a face image is invaluable to many conventional facial analysis problems mentioned above.
% as dense and fine as possible~\cite{cao2014displaced,cao20133d}. 

%realize the increasing importance and necessity of aligning the face images with a dense face shape.
%A dense face shape can provide a detailed representation of face which shows small details of face such as wrinkles and moles in a face. 
%Many facial problems require the face alignment algorithm to provide the shape as dense and fine as possible~\cite{cao2014displaced,cao20133d}. 
%We aim to look for an ideal model that is able to provide accurate dense face shape for any $2$D face images. 
%The estimated dense shape should provide correct estimation for both the fudicial facial points and the non-fudicial facial points. 

%The early works on face alignment were focusing on frontal view faces with few sparse landmarks but gradually by improving feature extraction, regression methods and more available datasets the accuracy of face alignment and number of landmarks increased. 
Since this interpretation has gone beyond the sparse set of landmarks, fitting a dense $3$D face model to the face image is a reasonable way to achieve DeFA.
In this work, we choose to develop the idea of fitting a dense $3$D face model to an image, where the model with thousands of vertexes makes it possible for face alignment to go very ``dense".
$3$D face model fitting is well studied in the seminal work of $3$D Morphorbal Model ($3$DMM)~\cite{blanz1999morphable}.
We see a recent surge when it is applied to problems such as large-pose face alignment~\cite{jourabloo2016large,face-alignment-across-large-poses-a-3d-solution}, $3$D reconstruction~\cite{booth20173d}, and face recognition~\cite{tran2016regressing}, especially using the convolutional neural network (CNN) architecture.

%Some recent face alignment methods focused on large pose face alignment and estimation of a dense $3$D shape of face which enhance face alignment methods to the dense model fitting. Considering the mentioned benefits of dense face alignment, in this work, we propose a set of constraints for the estimated $3$D shape and attempt to build a model that can densely fit $2$D face images with different variations such as pose, occlusion and lightings with more details. 

\begin{figure}[t!]
\begin{center}
\includegraphics[width=0.48\textwidth]{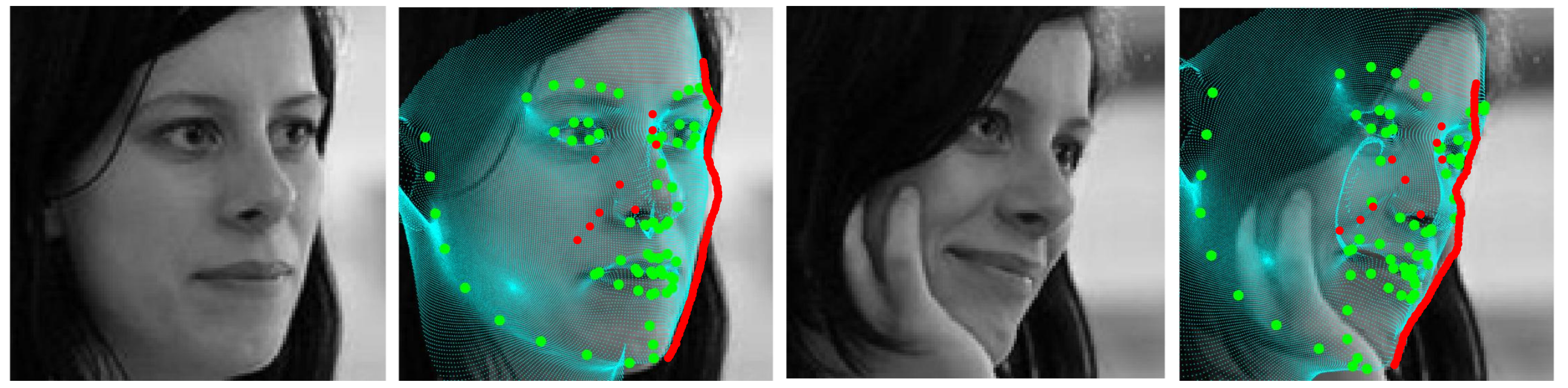} 
   \caption{A pair of images with their dense $3$D shapes obtained by imposing landmark fitting constraint, contour fitting constraint and sift pair constraint. }
\label{fig:FirstFig}\figvspace\vspace{-2mm}
\end{center}
\end{figure}   
However, most prior works on $3$D-model-fitting-based face alignment only utilize the sparse landmarks as supervision. 
There are two main challenges to be addressed in $3$D face model fitting, in order to enable high-quality DeFA.
%To tackle dense face alignment, the proposed model needs to solve two main issues. 
First of all, to the best of our knowledge, no public face dataset has dense face shape labeling. All of the in-the-wild face alignment datasets have no more than $68$ landmarks in the labeling.  
Apparently, to provide a high-quality alignment for face-region pixels, we need information more than just the landmark labeling.  
%To solve this, we gather texture information in the $2$D images such as contour, and SIFT feature, and we utilize them in the specifically designed loss functions.
Hence, the first challenge is to seek valuable information for additional supervision and integrate them in the learning framework.

Secondly, similar to many other data-driven problems and solutions, it is preferred that multiple datasets can be involved for solving face alignment task since a single dataset has limited types of variations. % and can not cover all of the challenging face alignment situations.
However, many face alignment methods can not leverage multiple datasets, because each dataset either is labeled differently. 
For instance, AFLW dataset~\cite{sagonas2013300} contains a significant variation of poses, but has a few number of visible landmarks. 
In contrast, $300$W dataset~\cite{sagonas2013300} contains a large number of faces with $68$ visible landmarks, but all faces are in a near-frontal view. %, lacking pose variations.
Therefore, the second challenge is to allow the proposed method to leverage multiple face datasets.
%In order to build a better model to handle above mentioned variations, multiple datasets should be used to include all possible variations. %Nevertheless, different datasets have different number of landmark annotations, such as $16$ landmark setting in Multi-PIE\cite{gross2010multi} and $68$ landmark setting in $300$-W~\cite{sagonas2013300}, which makes it hard to leverage different datasets in one single model. 
%Additionally, some unlabeled dataset also contain useful information, and can be utilized if process properly. 
%In this work, we also seek for leveraging multiple face datasets which have different number of landmarks to even unlabeled datasets.

%In previous method, XXX and XXX use 3DMM and provide dense model, but xxx.
With the objective of addressing both challenges, we learn a CNN to fit a $3$D face model to the face image.
While the proposed method works for any face image, we mainly pay attention to faces with large poses. Large-pose face alignment is a relatively new topic, and the performances in ~\cite{jourabloo2016large,face-alignment-across-large-poses-a-3d-solution} still have room to improve.
To tackle first challenge of limited landmark labeling, we propose to employ additional constraints. We include contour constraint where the contour of the predicted shape should match the detected $2$D face boundary, and SIFT constraint where the SIFT key points detected on two face images of the same individual should map to the same vertexes on the $3$D face model. 
Both constraints are integrated into the CNN training as additional loss function terms, where the end-to-end training results in an enhanced CNN for $3$D face model fitting.
For the second challenge of leveraging multiple datasets, the $3$D face model fitting approach has the inherent advantage in handling multiple training databases. 
Regardless of the landmark labeling number in a particular dataset, we can always define the corresponding $3$D vertexes to guide the training.
%Our method can leverage multiple datasets with different annotations to cover all possible variations, and adopt multiple constraints to supervise the learning in an end-to-end fashion. 
%There are a few $3$DMM-based face alignment such as~\cite{jourabloo2016large,zhu2016face} can provide a dense shape, but they cannot guarantee a dense fitting since the ground truth they provide is based on a limited set of landmarks.
%However, in this work, the proposed model can provide better fitting of face area other than just landmarks.

Generally, our main contributions can be summarized as:

1. We identify and define a new problem of dense face alignment, which seeks alignment of face-region pixels beyond the sparse set of landmarks.

2. To achieve dense face alignment, we develop a novel $3$D face model fitting algorithm that adopts multiple constraints and leverages multiple datasets.

3. Our dense face alignment algorithm outperforms the SOTA on challenging large-pose face alignment, and achieves competitive results on near-frontal face alignment. 
The model runs at real time.

%------------------------------------------------------------------------
\section{Related Work}

%------------------------------------------------------------------------

We review papers in three relevant areas: $3$D face alignment from a single image, using multiple constraints in face alignment, and using multiple datasets for face alignment.

%-------------------------------------------------------------------------
\Paragraph{$3$D model fitting in face alignment}
Recently, there are increasingly attentions in conducting face alignment by fitting the $3$D face model to the single $2$D image~\cite{jourabloo2016large,face-alignment-across-large-poses-a-3d-solution,liu2016joint,mcdonagh2016joint,zhao2016fast,pose-invariant-face-alignment-via-cnn-based-dense-3d-model-fitting}. 
%Wu \etal ~\cite{wu2016single} proposed a 3D interpreter network to estimate the 3D skeleton of a 2D object. The proposed model observes the constrained deformations of a certain 3D shape, and use projection parameters and shape deformations basis coefficients to describe the 3D reconstruction. 
In ~\cite{blanz1999morphable}, Blanz and Vetter proposed the $3$DMM to represent the shape and texture of a range of individuals. 
The analysis-by-synthesis based methods are utilized to fit the $3$DMM to the face image.
%After gathering and registering a data set of $3$D faces, they apply the Principal Component Analysis (PCA) to generate $3$D face bases and describe morphable model in the PCA space. They apply the $3$DMM model to the face recognition problem~\cite{blanz2003face}. 
In~\cite{face-alignment-across-large-poses-a-3d-solution,jourabloo2016large} a set of cascade CNN regressors with the extracted $3$D features is utilized to estimate the parameters of $3$DMM and the projection matrix directly. 
Liu \etal~\cite{liu2016joint} proposed to utilize two sets of regressors, for estimating update of $2$D landmarks and the other set estimate update of dense $3$D shape by using the $2$D landmarks update. 
They apply these two sets of regressors alternatively. 
Compared to prior work, our method imposes additional constraints, which is the key to dense face alignment.

%-------------------------------------------------------------------------
\Paragraph{Multiple constraints in face alignment}
Other than landmarks, there are other features that are useful to describe the shape of a face, such as contours, pose and face attributes. 
Unlike landmarks, those features are often not labeled in the datasets. 
Hence, the most crucial step of leveraging those features is to find the correspondence between the features and the $3$D shape.
In ~\cite{romdhani2005estimating}, multiple features constraints in the cost function is utilized to estimate the $3$D shape and texture of a $3$D face. 
$2$D edge is detected by Canny detector, and the corresponding $3$D edges' vertices are matched by Iterative Closest Point (ICP) to use this information. % to have more detailed $3$D shape.
%Similarly, Bas \etal ~\cite{bas2016fitting} use landmarks and edges to fit a $3$DMM model, and additionally discussed two ways to measure the correspondence between the ground truth edge and the predicted edge.
Furthermore, ~\cite{sanchez2016statistical} provides statistical analysis about the $2$D face contours and the $3$D face shape under different poses.

There is a few work using constraints as separate side tasks to facilitate face alignment.
In ~\cite{posealignmentBMVC}, they set a pose classification task, predicting faces as left, right profile or frontal, in order to assist face alignment. Even with such a rough pose estimation, this information boosts the alignment accuracy. 
Zhang \etal~\cite{zhang2016learning} jointly estimates $2$D landmarks update with the auxiliary attributes (e.g., gender, expression) in order to improve alignment accuracy. 
The ``mirrorability" constraint is used in~\cite{yang2015mirror} to force the estimated $2$D landmarks update be consistent between the image and its mirror image. 
In contrast, we integrate a set of constraints in an end-to-end trainable CNN to perform $3$D face alignment.

\begin{figure*}[t!]
\begin{center}
\includegraphics[scale = 0.35]{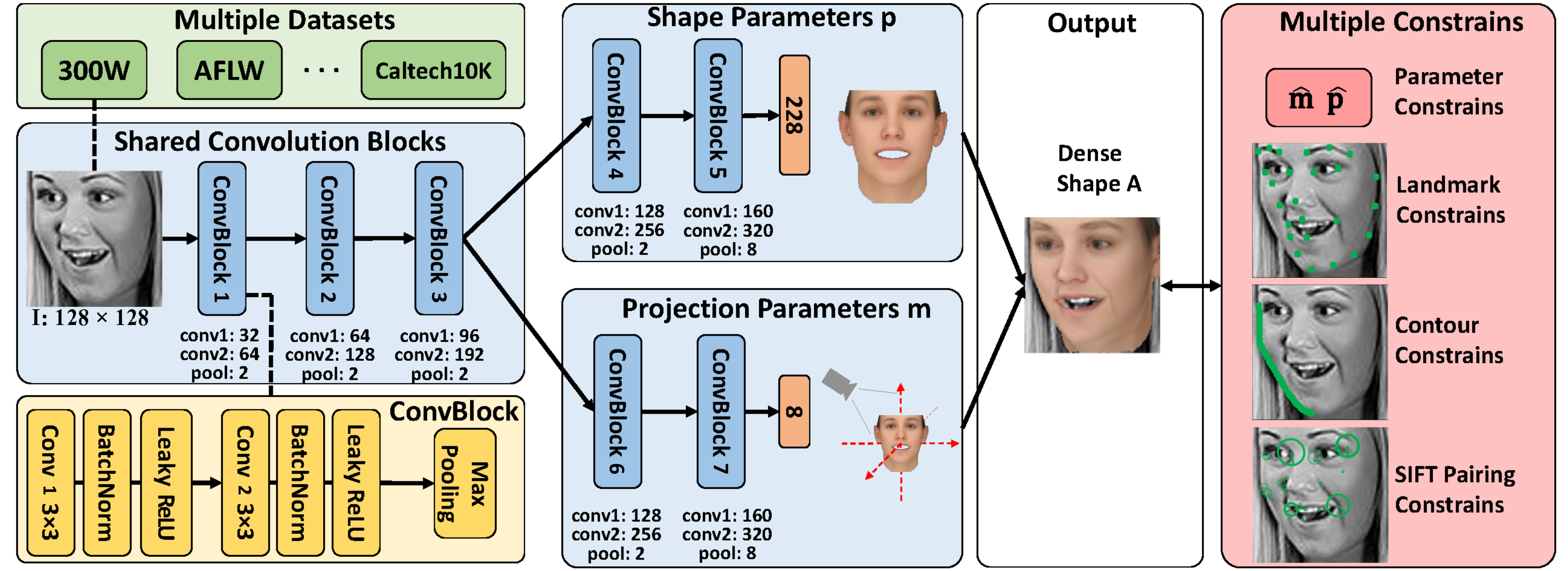}
   \caption{Architecture of CNN in the proposed DeFA method. The structure of each ConvBlock is shown in yellow area in the left bottom corner. Each convolution layer and fully connected layer is followed with one batch normalization layer (BN) and one leaky ReLU layer. The output dimension of each convolution layer is shown in the bottom of each unit, such as conv$1$: $32$, which means the output has $32$ channels. pool: $2$ denotes the pooling layer adopts a stride of $2$.}
\label{fig:CNNArchi}\figvspace
\end{center}
\end{figure*}

%-------------------------------------------------------------------------
\Paragraph{Multiple datasets in face alignment}
Despite the huge advantages (e.g., avoiding dataset bias), there are only a few face alignment works utilizing multiple datasets, owing to the difficulty of leveraging different types of face landmark labeling. 
Zhu \etal~\cite{zhu2014transferring} propose a transductive supervised descent method to transfer face annotation from a source dataset to a target dataset, and use both datasets for training.
\cite{smith2014collaborative} ensembles a non-parametric appearance model, shape model and graph matching to estimate the superset of the landmarks. 
Even though achieving good results, it suffers from high computation cost. 
Zhang \etal~\cite{zhang2015leveraging} propose a deep regression network for predicting the superset of landmarks. 
For each training sample, the sparse shape regression is adopted to generate the different types of landmark annotations.
In general, most of the mentioned prior work learn to map landmarks between two datasets, while our method can readily handle an arbitrary number of datasets since the dense $3$D face model can bridge the discrepancy of landmark definitions in various datasets.

%------------------------------------------------------------------------
%------------------------------------------------------------------------

\section{Dense Face Alignment}
In this section, we explain the details of the proposed dense face alignment method. 
We train a CNN for fitting the dense $3$D face shape to a single input face image. 
We utilize the dense $3$D shape representation to impose multiple constraints, e.g., landmark fitting constraint, contour fitting constraint and SIFT pairing constraint, to train such CNN.

\subsection{$3$D Face Representation}
We represent the dense $3$D shape of the face as, $\textbf{S}$, which contains the $3$D locations of $Q$ vertices, 
\begin{equation} \label{eq:S}
\textbf{S} = \begin{pmatrix}
  x_1 & x_2 & \cdots & x_Q \\
  y_1 & y_2 & \cdots & y_Q \\
  z_1 & z_2 & \cdots & z_Q
 \end{pmatrix}.
\end{equation}

To compute $\textbf{S}$ for a face, we follow the $3$DMM to represent it by a set of 3D shape bases,
\begin{equation} \label{eq:3DMM}
\textbf{S} = \bar{\textbf{S}} + \sum_{i=1}^{N_{id}} p_{id}^i \textbf{S}_{id}^i + \sum_{i=1}^{N_{exp}} p_{exp}^i \textbf{S}_{exp}^i,
\end{equation} 
where the face shape $\textbf{S}$ is the summation of the mean shape $\bar{\textbf{S}}$ and the weighted PCA shape bases $\textbf{S}_{id}$ and $\textbf{S}_{exp}$
with corresponding weights of $\textbf{p}_{id},\textbf{p}_{exp}$. 
In our work, we use $199$ shape bases $\textbf{S}_{id}^i,i=\{1,...,199\}$ for representing identification variances such as tall/short, light/heavy, and male/female, and $29$ shape bases $\textbf{S}_{exp}^i,i=\{1,...,29\}$ for representing expression variances such as mouth-opening, smile, kiss and etc. 
Each basis has $Q=53,215$ vertices, which are corresponding to vertices over all the other bases. 
The mean shape $\bar{\textbf{S}}$ and the identification bases $\textbf{S}_{id}$ are from Basel Face Model~\cite{paysan20093d}, and the expression bases $\textbf{S}_{exp}$ are from FaceWarehouse~\cite{cao2014facewarehouse}. 

A subset of $N$ vertices of the dense $3$D face $\textbf{U}$ corresponds to the location of $2$D landmarks on the image, 
\begin{equation} \label{eq:U}
\textbf{U} = \begin{pmatrix}
  u_1 & u_2 & \cdots & u_N \\
  v_1 & v_2 & \cdots & v_N
 \end{pmatrix}.
\end{equation}

By considering weak perspective projection, we can estimate the dense shape of a $2$D face based on the $3$D face shape. 
The projection matrix has $6$ degrees of freedom and can model changes w.r.t.~scale, rotation angles (pitch $\alpha$, yaw $\beta$, roll $\gamma$), and translations ($t_x$, $t_y$). 
The transformed dense face shape $\textbf{A} \in \mathbb{R}^{3\times Q}$ can be represented as,  
\begin{equation} \label{eq:weakpro}
\textbf{A} = \begin{bmatrix}
m_1 & m_2 & m_3 & m_4 \\
m_5 & m_6 & m_7 & m_8 \\
m_9 & m_{10} & m_{11} & m_{12} \\
\end{bmatrix}\  
\begin{bmatrix} 
    \textbf{S}\\
    \textbf{1}^\T
\end{bmatrix} 
\end{equation}
\begin{equation} \label{eq:opro}
\textbf{U} =  \textbf{Pr} \cdot \textbf{A},
\end{equation}
where $\textbf{A}$ can be orthographically projected onto $2$D plane to achieve $\textbf{U}$. 
Hence, z-coordinate translation ($m_{12}$) is out of our interest and assigned to be $0$. The orthographic projection can be denoted as matrix $\textbf{Pr} = \begin{bmatrix}
  1 & 0 & 0 \\
  0 & 1 & 0
 \end{bmatrix}$.

Given the properties of projection matrix, the normalized third row of the projection matrix can be represented as the outer product of normalized first two rows,
 \begin{equation} \label{eq:opro}
  [\bar{m}_9,\bar{m}_{10},\bar{m}_{11}]=[\bar{m}_1,\bar{m}_2,\bar{m}_3]\times 
  [\bar{m}_4,\bar{m}_5,\bar{m}_6].
\end{equation}

Therefore, the dense shape of an arbitrary $2$D face can be determined by the first two rows of the projection parameters $\textbf{m} = [m_1,\cdots,m_8]\in \mathbb{R}^8$ and the shape basis coefficients $\textbf{p}= [p_{id}^1,...,p_{id}^{199},p_{exp}^1,...p_{exp}^{29}]\in \mathbb{R}^{228}$.
The learning of the dense $3$D shape is turned into the learning of $\textbf{m}$ and $\textbf{p}$, which is much more manageable in term of the dimensionality.

%------------------------------------------------------------------------
\subsection{CNN Architecture}
%Due to the success of deep learning  methods in computer vision and non-linearity of mapping in our method, we utilize a CNN for learning the mapping from the input image $\textbf{I}$ to the projection parameters and the shape bases coefficients. 
Due to the success of deep learning in computer vision, we employ a convolutional neural network (CNN) to learn the nonlinear mapping function $f(\Theta)$ from the input image $\textbf{I}$ to the corresponding projection parameters $\textbf{m}$ and shape parameters $\textbf{p}$. 
The estimated parameters can then be utilized to construct the dense $3$D face shape.

Our CNN network has two branches, one for predicting $\textbf{m}$ and another for $\textbf{p}$, shown in Fig.~\ref{fig:CNNArchi}.
Two branches share the first three convolutional blocks. After the third block, we use two separate convolutional blocks to extract task-specific features, and two fully connected layers to transfer the features to the final output. 
Each convolutional block is a stack of two convolutional layers and one max pooling layer, and each conv/fc layer is followed by one batch normalization layer and one leaky ReLU layer. 

%This design can overcome the pathological curvature issue mention in ~\cite{face-alignment-across-large-poses-a-3d-solution}. Unlike ~\cite{face-alignment-across-large-poses-a-3d-solution,large-pose-face-alignment-via-cnn-based-dense-3d-model-fitting}, we employ one single CNN without any recursive loop, which makes the system capble to run in real-time. 

%The main loss function of the CNN is parameter loss which impose the parameter constraint (PC),
%which minimizes the difference of estimated parameters and the ground truth parameters.   

In order to improve the CNN learning, we employ a loss function including multiple constraints: Parameter Constraint (PC) $ J_{pr}$ minimizes the difference between the estimated parameters and the ground truth parameters; Landmark Fitting Constraint (LFC) $J_{lm}$ reduces the alignment error of $2$D landmarks;
 Contour Fitting Constraint (CFC) $J_{c}$ enforces the match between the contour of the estimated $3$D shape and the contour pixels of the input image; and SIFT Pairing Constraint (SPC) $J_{s}$ encourages that the SIFT feature point pairs of two face images to correspond to the same $3$D vertices.

We define the overall loss function as,
\begin{eqnarray} \label{eq4} %\nonumber \\
\argmin_{\hat{\textbf{m}},\hat{\textbf{p}}} J = J_{pr}+\lambda_{lm}J_{lm}
+\lambda_{c}J_{c}
+ \lambda_{s}J_{s},
\end{eqnarray}
where the parameter constraint (PC) loss is defined as,
\begin{equation} \label{equ:lm}
J_{pr} = \left\| \begin{bmatrix}   %\argmin_{\hat{\textbf{m}},\hat{\textbf{p}}} 
  \textbf{m}^\T \\
  \textbf{p}^\T 
 \end{bmatrix}  -  \begin{bmatrix}
  \hat{\textbf{m}}^\T \\
  \hat{\textbf{p}}^\T
 \end{bmatrix} \right\|^2.
\end{equation}

%------------------------------------------------------------------------
%\subsection{Landmark Fitting Constraint (LFC)} 
Landmark Fitting Constraint (LFC) aims to minimize the difference between the estimated $2$D landmarks and the ground truth $2$D landmark labeling $\textbf{U}_{lm} \in  \mathbb{R}^{2\times N}$. 
Given $2$D face images with a particular landmark labeling, we first manually mark the indexes of the $3$D face vertices that are anatomically corresponding to these landmarks. The collection of these indexes is denoted as $\textbf{i}_{lm}$. 
After the shape $\textbf{A}$ is computed from Eqn.~\ref{eq:weakpro} with the estimated $\hat{\textbf{m}}$ and $\hat{\textbf{p}}$, the $3$D landmarks can be extracted from $\textbf{A}$ by $\textbf{A} (:, \textbf{i}_{lm} )$. 
With projection of $\textbf{A} (:, \textbf{i}_{lm} )$ to $2$D plain, the LFC loss is defined as,
\begin{equation} \label{equ:lm}
 J_{lm} = \frac{1}{L} \cdot \Vert \textbf{Pr} \textbf{A} (:, \textbf{i}_{lm} ) - \textbf{U}_{lm} \Vert_F^2,
\end{equation}
where the subscript $F$ represents the Frobenius Norm, and $L$ is the number of pre-defined landmarks.
%Expanded formula of loss function derivatives is shown in the supplementary material.

%------------------------------------------------------------------------

\begin{figure}[t!]
\begin{tabular}{ccc}
\includegraphics[scale = 0.22]{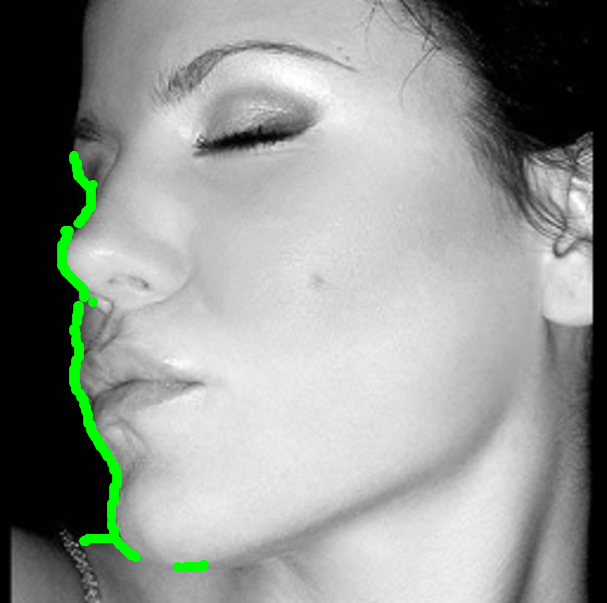}&
\includegraphics[scale = 0.22]{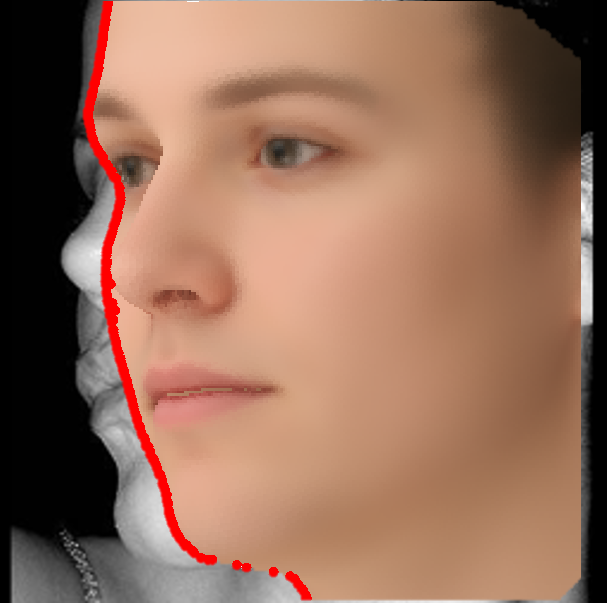}&
\includegraphics[scale = 0.22]{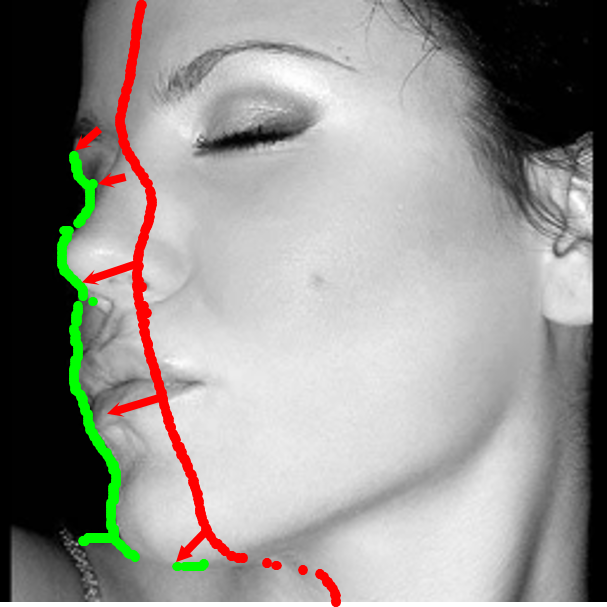}\\
(a)&(b)&(c)
\end{tabular}
   \caption{The CFC fitting process. $\textbf{A}_c$ is computed from estimated 3D face shape and $\textbf{U}_c$ is computed from the off-the-shelf edge detector. Contour correspondence is obtained via Closest Pair Algorithm, and loss $J_c$ is calculated based on Eqn.~\ref{equ:contour}}
\label{fig:contour}\figvspace
\end{figure}

\subsection{Contour Fitting Constraint (CFC)}
Contour Fitting Constraint (CFC) aims to minimize the error between the projected outer contour (i.e., silhouette) of the dense $3$D shape and the corresponding contour pixels in the input face image. 
The outer contour can be viewed as the boundary between the background and the $3$D face while rendering 3D space onto a $2$D plane. 
% and it shows the change of vertices' visibility from visible to invisible. Hereafter, we shortly call it as contour. 
On databases such as AFLW where there is a lack of labeled landmarks on the silhouette due to self-occlusion, this constraint can be extremely helpful.
%In such cases, landmark constraint only can lead to a highly inaccurate estimate which yet fit the visible landmarks properly, shown in Fig XXX.

To utilize this contour fitting constraint, we need to follow these three steps: 
1) Detect the true contour in the $2$D face image;
2) Describe the contour vertices on the estimated $3$D shape $\textbf{A}$; and
3) Determine the correspondence between true contour and the estimated one, and back-propagate the fitting error. 

First of all, we adopt an off-the-shelf edge detector, HED~\cite{xie2015holistically}, to detect the contour on the face image, $\textbf{U}_c \in \mathbb{R}^{2\times L}$. 
The HED has a high accuracy at detecting significant edges such as face contour in our case. 
Additionally, in certain datasets, such as $300$W~\cite{sagonas2013300} and AFLW-LPFA~\cite{jourabloo2016large}, additional landmark labelings on the contours are available. 
Thus we can further refine the detected edges by only retaining edges that are within a narrow band determined by those contour landmarks, shown in Fig~\ref{fig:contour}.a.
This preprocessing step is done offline before the training starts.

In the second step, the contour on the estimated $3$D shape $\textbf{A}$ can be described as the set of boundary vertices $\textbf{A} (:, \textbf{i}_{c} ) \in \mathbb{R}^{3\times L}$. 
$\textbf{A}$ is computed from the estimated $\hat{\textbf{m}}$ and $\hat{\textbf{p}}$ parameters. 
By utilizing the Delaunay triangulation to represent shape $\textbf{A}$, one edge of a triangle is defined as the boundary if the adjacent faces have a sign change in the $z$-values of the surface normals. 
This sign change indicates a change of visibility so that the edge can be considered as a boundary. 
The vertices associated with this edge are defined as boundary vertices, and their collection is denoted as $\textbf{i}_c$. % shows the indexes of selected edge vertices. %Additionally, we only extract contour from the $\alpha$-shape of $\textbf{A}$'s Delaunay triangulation. It will provide us a clean contour segment. 
This process is shown in Fig~\ref{fig:contour}.b.

In the third step, the point-to-point correspondences between $\textbf{U}_c$ and $\textbf{A} (:, \textbf{i}_{c} )$ are needed in order to evaluate the constraint. 
Given that we normally detect partial contour pixels on $2$D images while the contour of $3$D shape is typically complete, we match the contour pixel on the $2$D images with closest point on $3$D shape contour, and then calculate the minimun distance.
The sum of all minimum distances is the error of CFC, as shown in the Eqn.~\ref{equ:contour}.
To make CFC loss differentiable, we rewrite Eqn.~\ref{equ:contour} to compute the vertex index of the closest contour projection point, i.e., $k^0=\argmin_{k\in\textbf{i}_{c}}\Vert \textbf{Pr}\textbf{A}(:,k)-\textbf{U}_c(:,j)\Vert^2$.
Once $k^0$ is determined, the CFC loss will be differentiable, similar to Eqn.~\ref{equ:lm}.
%are assigned to the closest pairs of two points between the contour vertices and the edge pixels in the image. Assuming that the ground truth contour vertices are in a local neighborhood, the closest-pair approach can guarantee the convergence of Eqn.~\ref{equ:contour} to its minimum. Moreover, the assumption of the local neighborhood can be fulfilled by a partially-trained model that is able to make roughly correct estimation. 
%Overall landmark loss $J_{lm}$ of previous training epoch is set up as the trigger of whether to turn the contour fitting constraint on or off for current training epoch.

%Formally, the true contour in the image can be denoted as a set of $2$D points $\textbf{U}_c \in \mathbb{R}^{2\times L}$. 
%With all points in $\textbf{U}_c$ corresponded to vertices $\textbf{i}_{c}$, the error for CFC can be similarly generated by the following loss function,
\begin{multline} \label{equ:contour}
J_{c} = 
\frac{1}{L} \sum_j 
     \min_{k\in\textbf{i}_{c}}\Vert \textbf{Pr}\textbf{A}(:,k)-\textbf{U}_c(:,j)\Vert^2 \\
=\frac{1}{L} \sum_j \Vert
     \textbf{Pr}  \textbf{A} (:, \argmin_{k\in\textbf{i}_{c}}\Vert \textbf{Pr}\textbf{A}(:,k)-\textbf{U}_c(:,j)\Vert^2) \\
     - \textbf{U}_c (:,j) 
     \Vert^2.
\end{multline} 
Note that while $\textbf{i}_{c}$ depends on the current estimation of $\lbrace\textbf{m},\textbf{p}\rbrace$, for simplicity $\textbf{i}_{c}$ is treated as constant when performing back-propagation w.r.t.~$\lbrace\textbf{m},\textbf{p}\rbrace$. 
%In summary, Contour Fitting Constraint is implemented as Algorithm 1.

%------------------------------------------------------------------------
\subsection{SIFT Pairing Constraint (SPC)}
SIFT Pairing Constraint (SPC) regularizes the predictions of dense shape to be consistent on the significant facial points other than pre-defined landmarks, such as edges, wrinkles, and moles. 
The Scale-invariant feature transform (SIFT) descriptor is a classic  local representation that is invariant to image scaling, noise, and illumination. 
It is widely used in many regression-based face alignment methods~\cite{xiong2013supervised,tzimiropoulos2014gauss} to extract the local information. 

\begin{figure}[t!]
\includegraphics[scale = 0.28]{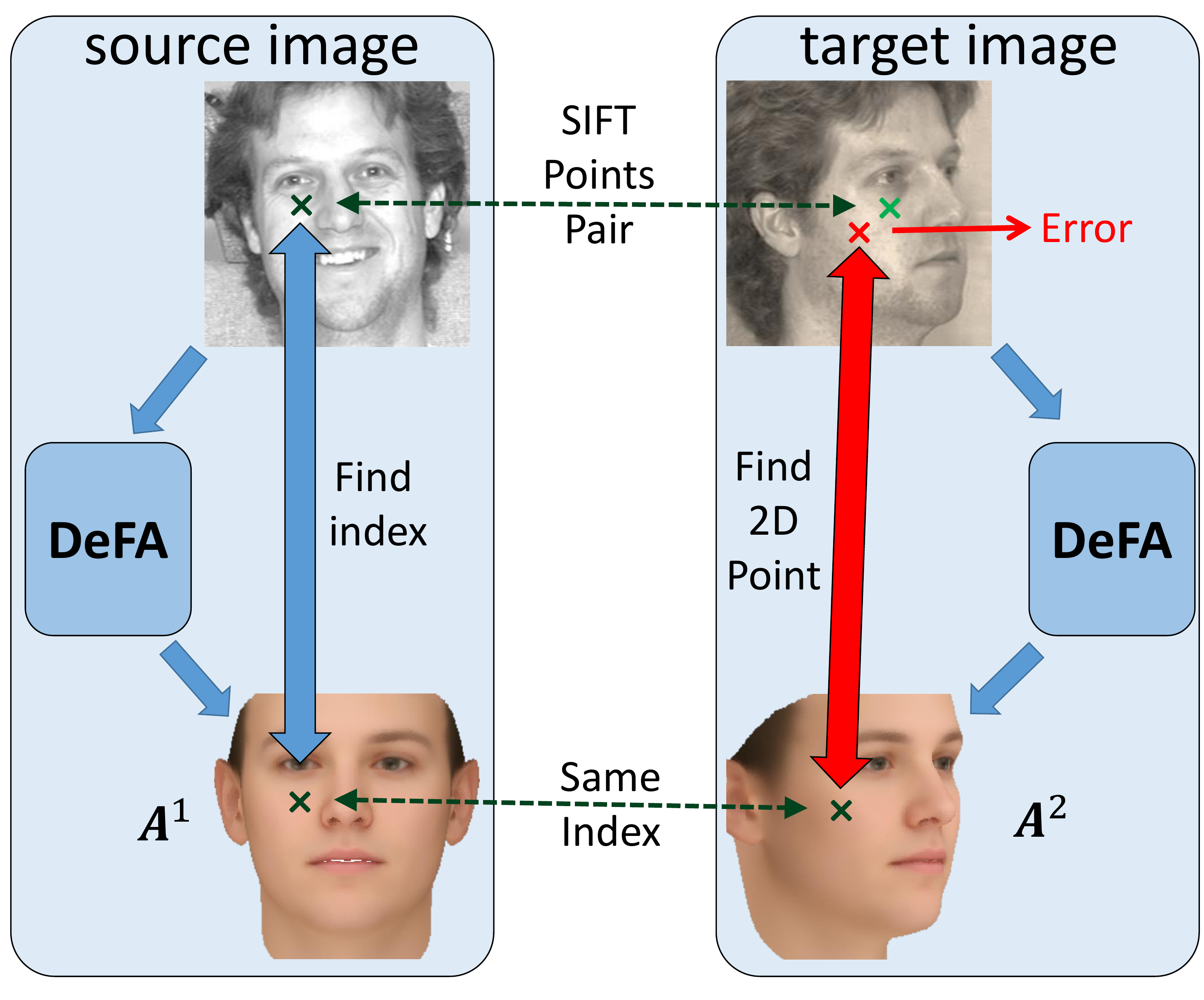}
   \caption{The illustration of the SIFT Matching process. }
\label{fig:sift}\figvspace
\end{figure}

In our work, the SIFT descriptors are used to detect and represent the significant points within the face pair. 
The face pair can either come from the same people with different poses and expressions, or the same image with different augmentation, e.g., cropping, rotation and $3$D augmentation, shown in Fig.~\ref{fig:sift}. 
The more face pairs we have, the stronger this constraint is. 
Given a pair of faces $i$ and $j$, we first detect and match SIFT points on two face images.
The matched SIFT points are denoted as $\textbf{U}_s^{i}$ and $\textbf{U}_s^{j} \in \mathbb{R}^{2\times L_{ij}}$.
%Instead, we can set up the correspondence between faces, and the SIFT points can be matched among the different face images. 

With a perfect dense face alignment, the matched SIFT points would overlay with exactly the same vertex in the estimated $3$D face shapes, denoted as $\textbf{A}^i$ and $\textbf{A}^j$.
In practices, to verify how likely this ideal world is true and leverage it as a constraint, we first find the $3$D vertices $\textbf{i}_s^i$ whose projections overlay with the $2$D SIFT points, $\textbf{U}_s^{i}$.
\begin{equation} \label{sift}
\textbf{i}_s^i = \arg\min_{i\in \{ 1,...,L_{ij} \}} \Vert \textbf{A}^i\{\textbf{i}_s^{i}\} - \textbf{U}_s^{i} \Vert_F^2,
\end{equation}

Similarly, we find $\textbf{j}_s^j$ based on $\textbf{U}_s^{j}$.  
Now we define the SPC loss function as
\begin{multline} \label{js}
%(\textbf{m}^i,\textbf{p}^i,\textbf{m}^j,\textbf{p}^j)
 J_{s}(\hat{\textbf{m}}^j,\hat{\textbf{p}}^j,\hat{\textbf{m}}^i,\hat{\textbf{p}}^i) = \\
\frac{1}{L_{ij}} \left( \Vert \textbf{A}^i\{\textbf{i}_s^{j}\} - \textbf{U}_s^{i} \Vert_F^2 + 
\Vert \textbf{A}^j\{\textbf{i}_s^{i}\} - \textbf{U}_s^{j} \Vert_F^2 \right)
\end{multline}
where $\textbf{A}^i$ is computed using $\lbrace\textbf{m}^i,\textbf{p}^i\rbrace$. 
As shown in Fig.~\ref{fig:sift}, we map SIFT points from one face to the other and compute their distances w.r.t.~the matched SIFT points on the other face. 
With the mapping from both images, we have two terms in the loss function of Eqn.~\ref{js}.
%and $ \textbf{i}_s^{i} = \lbrace i_1,i_2,..., i_{L_{ij}} \rbrace$ denotes the vertices of SIFT pairing points on image $i$, $ \textbf{i}_s^{i}$ satisfies Eqn.~\ref{sift}.

%------------------------------------------------------------------------

\section{Experimental Results}
%------------------------------------------------------------------------

\subsection{Datasets}

We evaluate our proposed method on four benchmark datasets: AFLW-LFPA~\cite{jourabloo2015pose}, AFLW$2000$-$3$D~\cite{face-alignment-across-large-poses-a-3d-solution}, $300$W~\cite{sagonas2013300} and IJBA~\cite{klare2015pushing}. 
All datasets  used in our training and testing phases are listed in Tab.~\ref{table:dataset}. 

\noindent
$\textbf{AFLW-LFPA:}$ AFLW contains around $25,000$ face images with yaw angles between $\pm90^{\circ}$, and each image is labeled with up to $21$ visible landmarks. 
In~\cite{jourabloo2015pose}, a subset of AFLW with a balanced distribution of the yaw angle is introduced as AFLW-LFPA. 
It consists of $3,901$ training images and $1,299$ testing images. 
Each image is labeled with $13$ additional landmarks. 

\noindent
$\textbf{AFLW2000-3D:}$ Prepared by~\cite{face-alignment-across-large-poses-a-3d-solution}, this dataset contains $2,000$ images with yaw angles between $\pm90^{\circ}$ of the AFLW dataset. 
Each image is labeled with $68$ landmarks.
Both this dataset and  AFLW-LFPA are widely used for evaluating large-pose face alignment.

\noindent
$\textbf{IJBA:}$ IARPA Janus Benchmark A (IJB-A)~\cite{klare2015pushing} is an in-the-wild dataset containing $500$ subjects and $25,795$ images with three landmark, two landmarks at eye centers and one on the nose.
While this dataset is mainly used for face recognition, the large dataset size and the challenging variations (e.g., $\pm90^{\circ}$ yaw and images resolution) make it suitable for evaluating face alignment as well.

\noindent
$\textbf{300W:}$ $300$W~\cite{sagonas2013300} integrates multiple databases with standard $68$ landmark labels, including AFW~\cite{zhu2012face}, LFPW~\cite{belhumeur2013localizing}, HELEN~\cite{zhou2013extensive}, and IBUG~\cite{sagonas2013300}. 
This is the widely used database for evaluating near-frontal face alignment.

\noindent
$\textbf{COFW~\cite{burgos2013robust}:}$ This dataset includes near-frontal face images with occlusion. 
%Each image is labeled with $29$ landmarks and visibilty label.
We use this dataset in training to make the model more robust to occlusion.

\noindent
$\textbf{Caltech10k~\cite{angelova2005pruning}:}$ It contains four labeled landmarks: two on eye centers, one on the top of the nose and one mouth center. We do not use the mouth center landmark since there is no corresponding vertex  on the $3$D shape existing for it.

\noindent
$\textbf{LFW~\cite{learned2016labeled}:}$ Despite having no landmark labels, LFW can be used to evaluate how dense face alignment method performs via the corresponding SIFT points between two images of the same individual.

% training set and follow the data augmentation method in ~\cite{face-alignment-across-large-poses-a-3d-solution} to generate large-pose training face images ($96,268$ images) and 

\begin{table}[t!]
	\caption{The list of face datasets used for training and testing. %$^{\textbf{*}}$: Caltech10K originally has $5$ landmark labelings. However, the fifth landmark is in the center of the mouth, which isn't corresponding to any shape vertex when the mouth is open. Hence we exclude the fifth landmark. $^{\textbf{**}}$: Due to large pose, landmarks in AFLW-LFPA are up to 34 in each image.
	} % title of Table
      \label{table:dataset}
      \centering
        \begin{tabular}{c c c c} % centered columns (4 columns)
              \hline %inserts double horizontal lines
              Database & Landmark & Pose & Images\\[0.5ex] 
              \hline\hline
              \multicolumn{4}{|c|}{$Training$} \\
              \hline 
              300W~\cite{sagonas2013300}       & $68$ & Near-frontal & $3,148$\\
              300W-LP~\cite{face-alignment-across-large-poses-a-3d-solution}    & $68$ & $[-90^\circ,90^\circ]$ & $96,268$\\
              Caltech10k~\cite{angelova2005pruning} & $4$ & Near-frontal &$10,524$\\
              AFLW-LFPA~\cite{jourabloo2015pose} & $21$  & $[-90^\circ,90^\circ]$ & $3,901$\\
              COFW~\cite{burgos2013robust}     & $29$ & Near-frontal & $1,007$ \\
              \hline
              \multicolumn{4}{|c|}{$Testing$} \\
              \hline 
              AFLW-LFPA~\cite{jourabloo2015pose}  & $34$ & $[-90^\circ,90^\circ]$ & $1,299$\\
              AFLW2000-3D~\cite{face-alignment-across-large-poses-a-3d-solution}& $68$ & $[-90^\circ,90^\circ]$ & $2,000$\\
              300W~\cite{sagonas2013300} & $68$  & Near-frontal & $689$\\
              IJB-A~\cite{klare2015pushing}    & $3$ & $[-90^\circ,90^\circ]$ & $25,795$ \\
              LFW~\cite{learned2016labeled}      & $0$ & Near-frontal & $34,356$ \\
              \hline\hline %inserts single line
        \end{tabular}
\end{table}

\vspace{-1mm}
%------------------------------------------------------------------------
\subsection{Experimental setup}
\vspace{-2mm}
\noindent
$\textbf{Training sets and procedures}:$ While utilizing multiple datasets is beneficial for learning an effective model, it also poses challenges to the training procedure.
To make the training more manageable, we train our DeFA model in three stages, with the intention to gradually increase the datasets and employed constraints.
At stage $1$, we use $300$W-LP to train our DeFA network with parameter constraint (PL). 
At stage $2$, we additionally include samples from the  Caltech$10$K~\cite{angelova2005pruning}, and COFW~\cite{burgos2013robust}  to continue the training of our network with the additional landmark fitting constraint (LFC). 
At stage $3$, we fine-tune the model with SPC and CFC constraints. 
For large-pose face alignment, we fine-tune the model with AFLW-LFPA training set.  
For near-frontal face alignment, we fine-tune the model with $300$W training set. 
All samples at the third stage are augmented $20$ times with up to $\pm 20^{\circ}$ random in-plain rotation and $15\%$ random noise on the center, width, and length of the initial bounding box. 
Tab.~\ref{table:stages} shows the datasets and constraints that are used at each stage.

\begin{table}[t!]
      \centering
        \caption{The list of datasets used in each training stage, and the employed constraints for each dataset: Parameter Constraint (PC); Landmark Fitting Constraint (LFC); SIFT Pairing Constraint (SPC); Contour Fitting Constraint (CFC). }
         \begin{tabular}{c c c c} % centered columns (4 columns)
              \hline %inserts double horizontal lines
              Dataset & Stage 1 & Stage 2 & Stage 3\\ [0.5ex] 
              \hline 
              300W-LP~\cite{face-alignment-across-large-poses-a-3d-solution}    & PC & 
              \begin{tabular}{c}
              PC\\[-1mm]
              LFC 
              \end{tabular} & -\\
              \hline 
              Caltech10k~\cite{angelova2005pruning} & - & LFC & -\\
              \hline 
              COFW~\cite{burgos2013robust}     & - & LFC & - \\
              \hline 
              AFLW-LFPA~\cite{jourabloo2015pose} & - & - & 
              \begin{tabular}{c}
              LFC\\[-1mm]
              SPC\\[-1mm]
              CFC
              \end{tabular}
              \\
              \hline 
              300W~\cite{sagonas2013300}       & - & - & 
              \begin{tabular}{c}
              LFC\\[-1mm]
              SPC\\[-1mm]
              CFC 
              \end{tabular}
              \\            
              \hline\hline
        \end{tabular}
        \label{table:stages}
\end{table}

%\vspace{-1mm}

\noindent 
$\textbf{Implementation details}$: Our DeFA model is implemented with MatConvNet~\cite{vedaldi15matconvnet}. 
To train the network, we use $20$, $10$, and $10$ epochs for stage $1$ to $3$. 
We set the initial global learning rate as $1e-3$, and reduce the learning rate by a factor of $10$ when the training error approaches a plateau. 
The minibatch size is $32$, weight decay is $0.005$, and the leak factor for Leaky ReLU is $0.01$. 
In stage 2, the regularization weights $\lambda_{pr}$ for PC is $1$ and $\lambda_{lm}$ for LFC is $5$; In stage 3, the regularization weights $\lambda_{lm}$, $\lambda_{s}$, $\lambda_{c}$ for LFC, SPC and CFC are set as $5$, $1$ and $1$, respectively.

%------------------------------------------------------------------------
\noindent 
$\textbf{Evaluation metrics}$: For performance evaluation and comparison, we use two metrics for normalizing the MSE.
 We follow the normalization method in~\cite{jourabloo2016large} for large-pose faces, which normalizes the MSE by using the bounding box size. 
We term this metric as ``NME-lp".
For the near-frontal view datasets such as $300$W, we use the  inter-ocular distance for normalizing the MSE, termed  as ``NME-nf".

\subsection{Experiments on Large-pose Datasets}
To evaluate the algorithm on large-pose datasets, we use the AFLW-LFPA, AFLW$2000$-$3$D, and IJB-A datasets. 
The results are presented in Tab.~\ref{table:large-pose}, where the performance of the baseline methods is either reported from the published papers or by running the publicly available source code.
%We show the reported performance in their papers for the compared methods. 
For AFLW-LFPA, our method outperforms the best methods with a large margin of $17.8\%$ improvement. 
For AFLW$2000$-$3$D, our method also shows a large improvement. Specifically, for images with yaw angle in $[60^{\circ},90^{\circ}]$, our method improves the performance by $28\%$ (from $7.93$ to $5.68$).
For the IJB-A dataset, %we use the available source code of the PAWF method to compare with our method. 
even though we are able to only compare the accuracy for the three labeled landmarks, our method still reaches a higher accuracy. 
Note that the best performing baselines, $3$DDFA and PAWF, share the similar overall approach in estimating $\mathbf{m}$ and $\mathbf{p}$, and also aim for large-pose face alignment.
The consistently superior performance of our DeFA indicates that we have advanced the state of the art in large-pose face alignment.

\begin{table*}[t]
\caption{The benchmark comparison (NME-lp) on three large-pose face alignment datasets.} % title of Table
\centering % used for centering table
\begin{tabular}{c c c c c c c} % centered columns (4 columns)
\hline %inserts double horizontal lines
Baseline & CFSS~\cite{zhu2015face} & PIFA~\cite{jourabloo2015pose} & CCL~\cite{zhu2016unconstrained} & 3DDFA~\cite{zhu2016face} & PAWF~\cite{jourabloo2016large} & $\textbf{Ours}$ \\ [0.5ex] % inserts table
%heading
\hline % inserts single horizontal line
AFLW-LFPA  & $6.75$ & $6.52$ & $5.81$ & - & $4.72$ & $\textbf{3.86}$\\ % inserting body of the table
AFLW$2000$-$3$D & -      & -      & -      & $5.42$ & - & $\textbf{4.50}$\\ % inserting body of the table
IJB-A & - & - & - & - & $6.76$ & $\textbf{6.03}$\\
\hline\hline %inserts single line
\end{tabular}
\label{table:large-pose}\figvspace
\end{table*}

\subsection{Experiments on Near-frontal Datasets}
Even though the proposed method can handle large-pose alignment, to show its performance on the near-frontal datasets, we evaluate our method on the $300$W dataset. 
The result of the state-of-the-art method on the both common and challenging sets are shown in Tab.~\ref{table:300W}. 
To find the corresponding landmarks on the cheek, we apply the landmark marching~\cite{zhu2015high} algorithm to move contour landmarks from self-occluded location to the silhouette.
Our method is the second best method on the challenging set. 
In general, the performance of our method is comparable to other methods that are designed for near-frontal datasets, especially under the following consideration.
That is, most prior face alignment methods do not employ shape constraints such as $3$DMM, which could be an advantage for near-frontal face alignment, but might be a disadvantage for large-pose face alignment.
The only exception in Tab.~\ref{table:300W} in $3$DDFA~\cite{face-alignment-across-large-poses-a-3d-solution}, which attempted to overcome the shape constraint by using the additional SDM-based finetuning. 
It is a strong testimony of our model in that DeFA, without further fine-tuning, outperforms both $3$DDFA and its fine tuned version with SDM.
%It achieves better performance than the $3$DDFA~\cite{face-alignment-across-large-poses-a-3d-solution} method, which similarly estimates $\textbf{m}$ and $\textbf{p}$ parameters, and the the combination of the $3$DDFA and SDM~\cite{xiong2013supervised} methods on both the common and the challenging sets.   
\begin{table}[t]
\caption{The benchmark comparison (NME-nf) on $300$W dataset. The top two performances are in bold.} % title of Table
	\resizebox{0.48\textwidth}{!} 
{

\centering % used for centering table
\begin{tabular}{c c c c} % centered columns (4 columns)
\hline %inserts double horizontal lines
Method & Common set & Challenging set & Full set \\ [0.5ex] % inserts table
%heading
\hline % inserts single horizontal line
RCPR~\cite{burgos2013robust} & $6.18$ & $17.26$ & $7.58$\\ 
SDM~\cite{xiong2013supervised}  & $5.57$ & $15.40$ & $7.50$\\ 
LBF~\cite{ren2014face}  & $4.95$ & $11.98$ & $6.32$\\ 
CFSS~\cite{zhu2015face} & $\textbf{4.73}$ & $9.98$ & $\textbf{5.76}$\\ 
RAR~\cite{xiao2016robust}  & $\textbf{4.12}$ & $\textbf{8.35}$ & $\textbf{4.94}$\\ 
\hline
3DDFA~\cite{face-alignment-across-large-poses-a-3d-solution}& $6.15$ & $10.59$ & $7.01$\\ 
3DDFA+SDM& $5.53$ & $9.56$ & $6.31$\\ 
$\textbf{DeFA}$  & $5.37$ & $\textbf{9.38}$ & $6.10$\\ 
\hline\hline %inserts single line
\end{tabular}
}
\label{table:300W}\figvspace
\end{table}

%------------------------------------------------------------------------
\subsection{Ablation Study}
To analyze the effectiveness of the DeFA method, we design two studies to compare the influence of each part in the DeFA and the improvement by adding each dataset.

Tab.~\ref{table:eval1} shows the consistent improvement achieved by utilizing more datasets in different stages and constraints according to Tab.~\ref{table:stages} on both testing datasets. 
It shows the advantage and the ability of our method in leveraging more datasets. 
The accuracy of our method on the AFLW$2000$-$3$D consistently improves by adding more datasets. For the AFLW-PIFA dataset, our method achieves $9.5\%$ and $20\%$ relative improvement by utilizing the datasets in the stage $2$ and stage $3$ over the first stage, respectively. 
If including the datasets from both the second and third stages, we can have $26\%$ relative improvement and achieve NME of $3.86\%$. 
Comparing the second and third rows in Tab.~\ref{table:eval1} shows that the effectiveness of  CFC and SPC is more than LFC. 
This is due to the utilization of more facial matching in the CFC and SPC.  
\begin{table}[t!]
\caption{The NME-lp when utilizing more datasets.} % title of Table
	\resizebox{0.48\textwidth}{!} 
{
\centering % used for centering table
\begin{tabular}{c c c c} % centered columns (4 columns)
\hline %inserts double horizontal lines
Training Stages & AFLW$2000$-$3$D &AFLW-LFPA \\ [0.5ex]
\hline
stage1 & $6.23$ & $5.24$\\ 
stage1 + stage2  & $5.68$ & $4.74$\\ 
stage1 + stage3  & $4.85$ & $4.15$\\ 
\hline
$\textbf{stage1 + stage2 + stage3}$  & $\textbf{4.50}$ & $\textbf{3.86}$\\ 
\hline\hline %inserts single line
\end{tabular}
}
\label{table:eval1}\figvspace
\end{table}

\begin{figure}[t]
\begin{center}
\begin{tabular}{c c}
\includegraphics[width=0.22\textwidth]{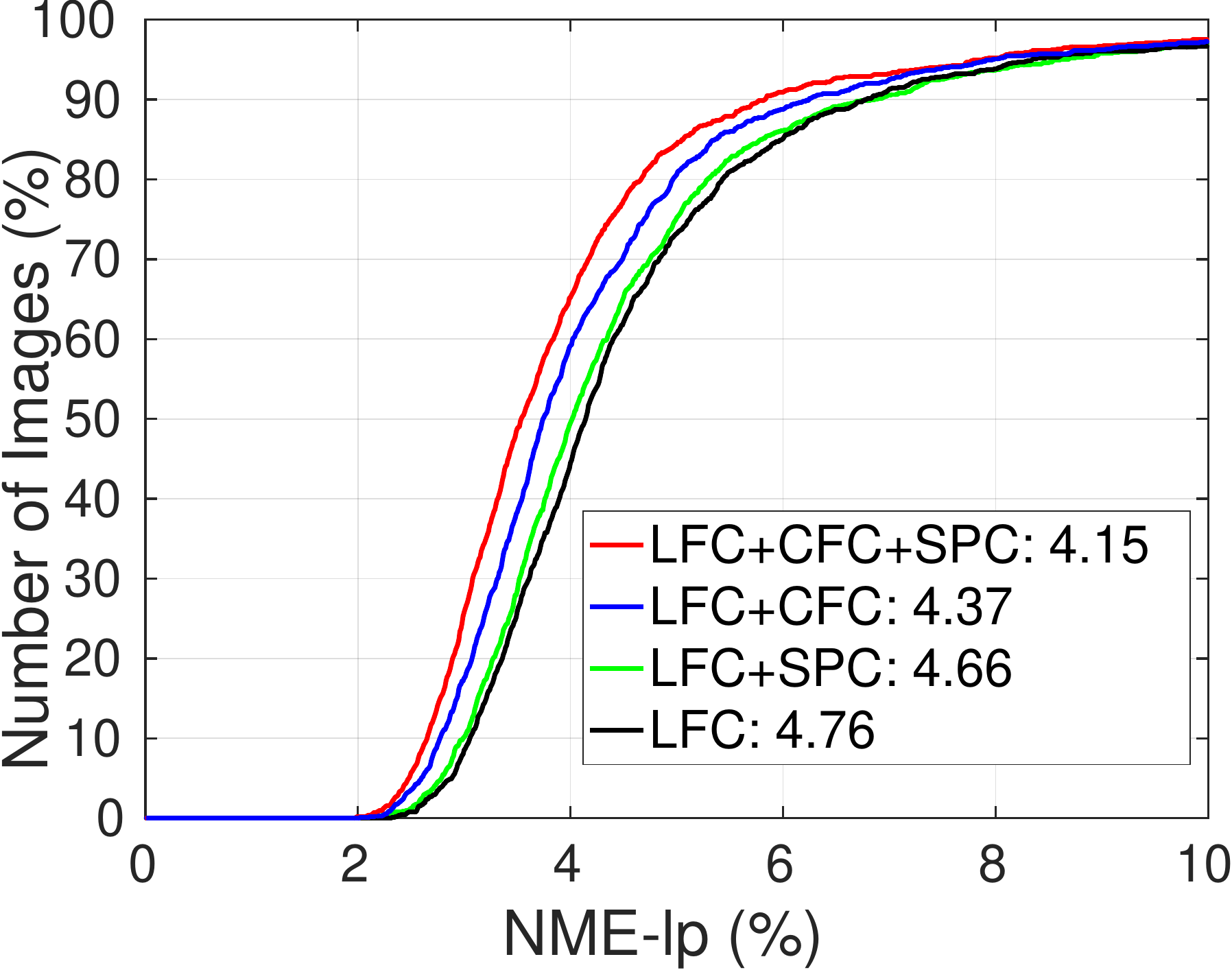} &
\includegraphics[width=0.22\textwidth]{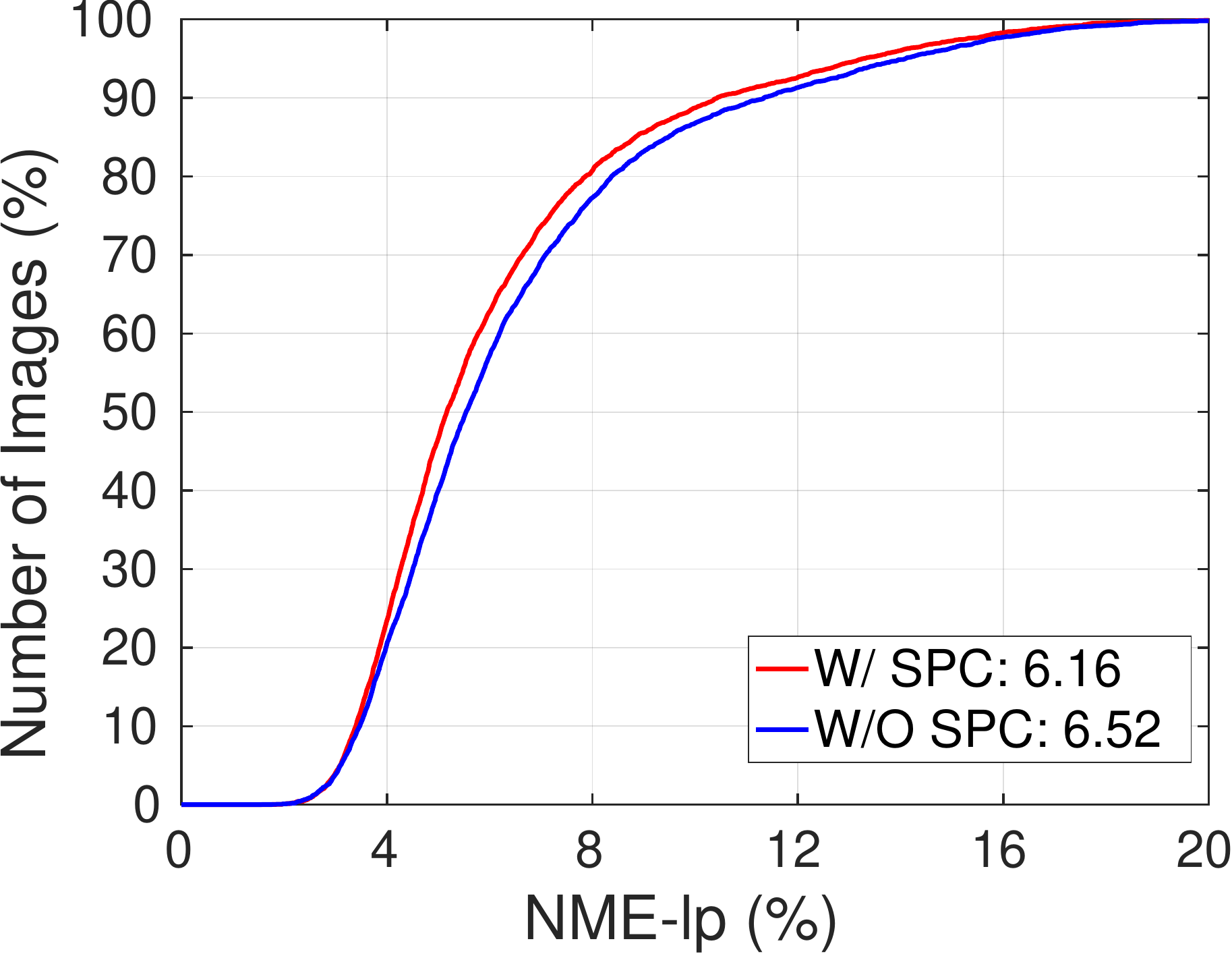} 
\end{tabular}
   \caption{Left: The effect of constraints in enhancing the accuracy on the AFLW-LPFA testing set. The  NME-lp of each setting is shown in legend. Right: The influence of the SIFT pairing constraint (SPC) in improving the performance for selected $5,000$ pairs from IJB-A.}
\label{fig:SPC}
\end{center}\figvspace
\end{figure}

The second study shows the performance improvement achieved by using the proposed constraints. 
%aims to show the influence of the proposed constraints.
We train models with different types of active constraints and test them on the AFLW-PIFA test set. 
 % the improvement achieves by activating each constraint one by one. 
Due to the time constraint, for this experiment, we did not apply $20$ times augmentation of the third stage's dataset. 
We show the results in the left of Fig.~\ref{fig:SPC}.
Comparing LFC+SPC and LFC+CFC performances shows that the CFC is more helpful than the SPC. 
The reason is that CFC is more helpful in correcting the pose of the face and leads to more landmark error reduction. 
Using all constraints achieves the best performance.

Finally, to evaluate the influence of using the SIFT pairing constraint (SPC), we use all of the three stages datasets to train our method. 
We select $5,000$ pairs of images from the IJB-A dataset and compute the NME-lp of all matched SIFT points according to Eqn.~\ref{js}. 
The right plot in Fig.~\ref{fig:SPC} illustrates the CED diagrams of NME-lp, for the trained models with and without the SIFT pairing constraint. 
This result shows  that for the images with NME-lp between $5\%$ and $15\%$ the SPC is helpful.

\begin{figure*}[t]
\begin{center}
\includegraphics[width=\textwidth]{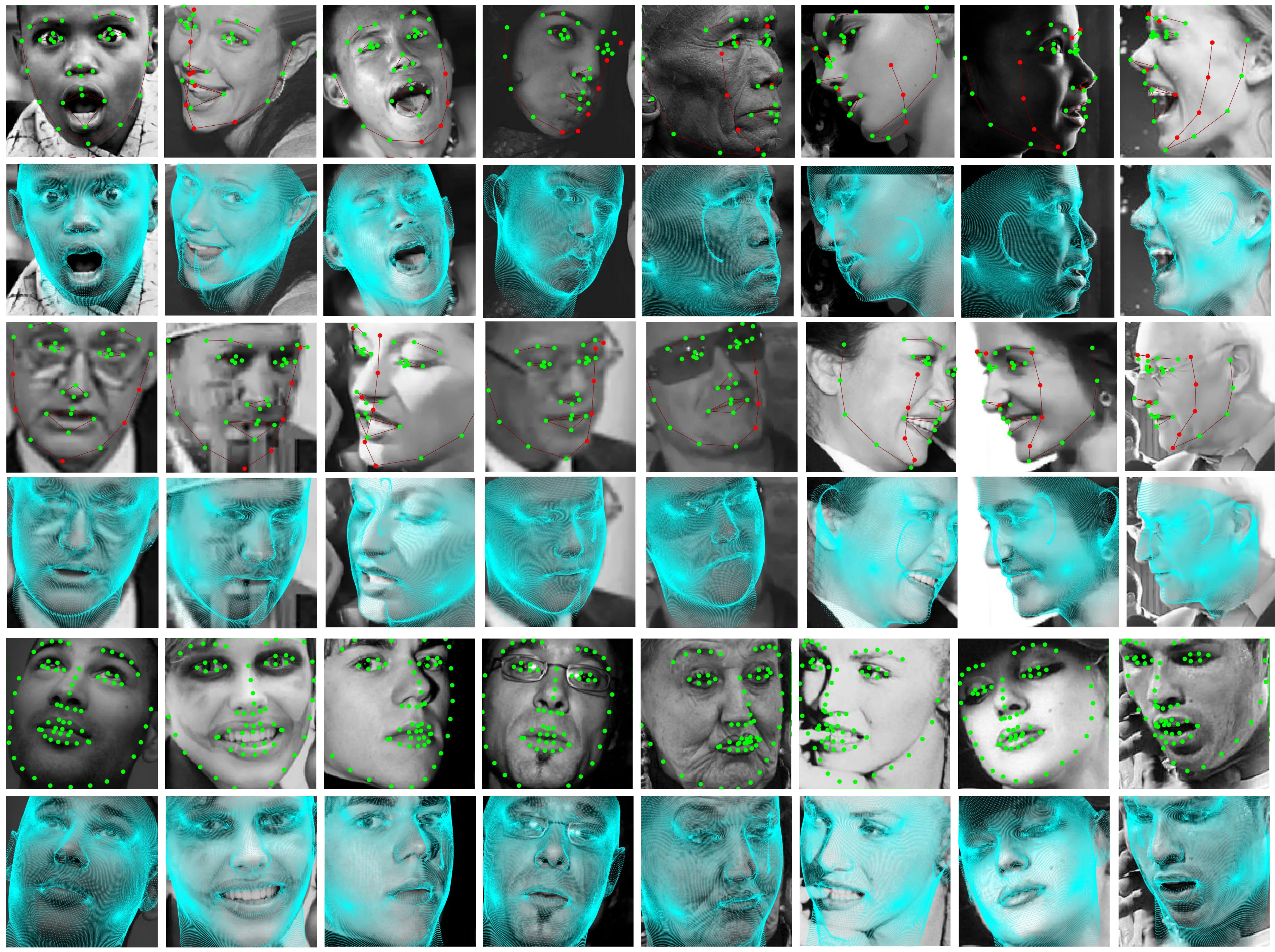} 
   \caption{The estimated dense $3$D shape and their landmarks with visibility labels for different datasets. From top to bottom, the results on AFLW-LPFA, IJB-A and $300$W datasets are shown in two rows each. The green landmark are visible and the red landmarks show the estimated locations for invisible landmarks. Our model can fit to diverse poses, resolutions, and expressions.}
\label{fig:LastFig}
\end{center}
\end{figure*}

\begin{figure}[t]
\begin{center}
\includegraphics[width=0.4\textwidth]{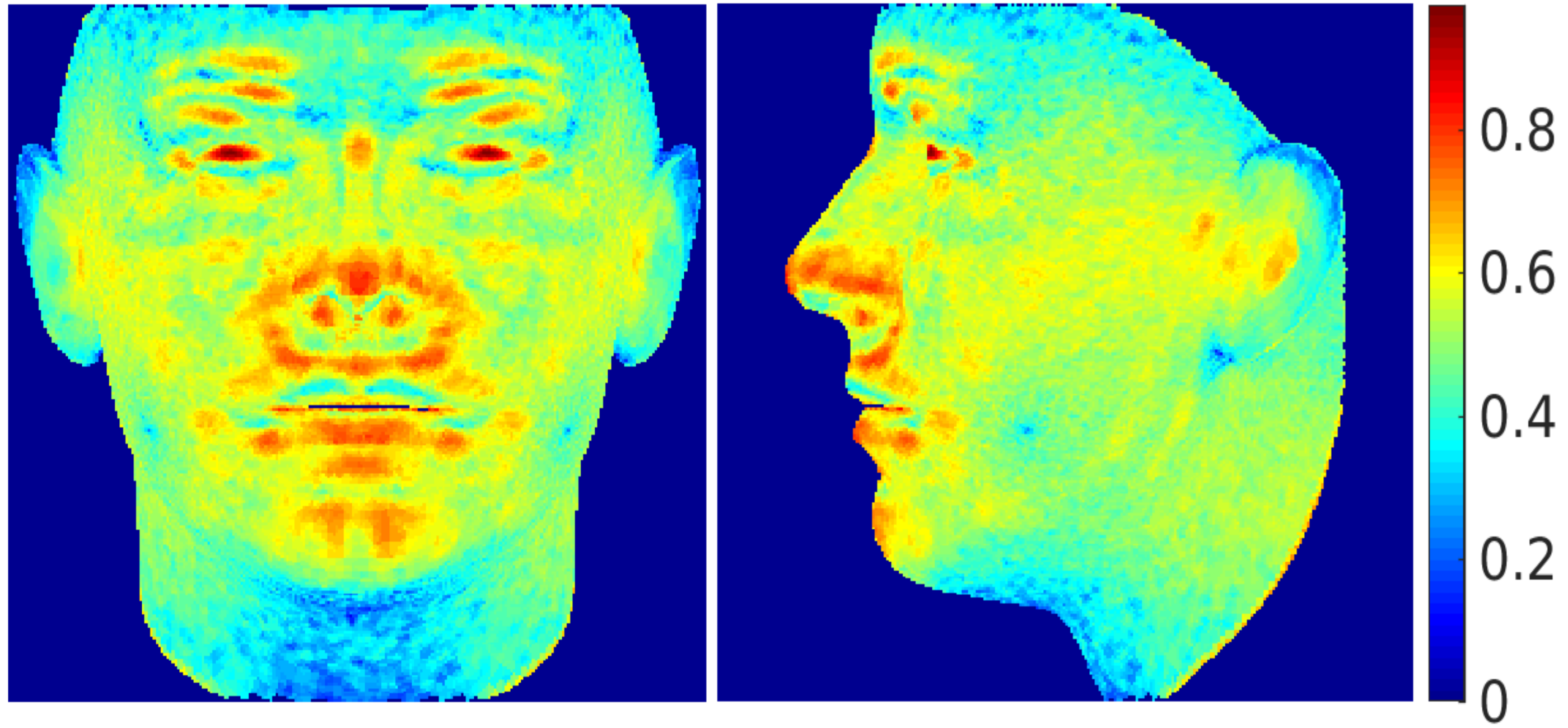} 
   \caption{The log plot of the number of matched SIFT points in the $300$W-LP training set. It shows that the SIFT constraints cover the whole face, especially the highly textured area.}
\label{fig:siftPoints} \figvspace
\end{center}
\end{figure}

Part of the reason DeFA works well is that it receives ``dense" supervision. 
%In order to show that our method favors from the information of the whole face during the training,
To show that, we take all matched SIFT points in the $300$W-LP dataset, find their corresponding vertices, and plot the log of the number of SIFT points on each of the $3$D face vertex.
 %we show the the normalized number of utilization of each vertex 
As shown in Fig.~\ref{fig:siftPoints}, SPC utilizes SIFT points to cover the whole $3$D shape and the points in the highly textured areas are substantially used.
We can expect that these SIFT constraints will act like anchors to guild the learning of the model fitting process. 

\section{Conclusion}
We propose a large-pose face alignment method which estimates accurate $3$D face shapes by utilizing a deep neural network. In addition to facial landmark fitting, we propose to align contours and the SIFT feature point pairs to extend the fitting beyond facial landmarks. Our method is able to leverage from utilizing multiple datasets with different landmark markups and numbers of landmarks. We achieve the state-of-the-art performance on three challenging large pose datasets and competitive performance on hard medium pose datasets.

\end{document}